\pdfoutput=1

\documentclass[11pt]{article}

\usepackage[]{ACL2023}

\usepackage{times}
\usepackage{latexsym}
\usepackage{amsmath}

\usepackage[T1]{fontenc}
\usepackage{makecell}

\usepackage[utf8]{inputenc}
\usepackage{booktabs}
\usepackage{graphicx}
\usepackage{adjustbox}
\usepackage{microtype}
\usepackage{arydshln}

\usepackage{inconsolata}

\usepackage{bm} 

\title{HINT: Hypernetwork Instruction Tuning \\ for Efficient Zero- \& Few-Shot Generalisation}

\author{Hamish Ivison $^\alpha$ \And
  Akshita Bhagia $^\alpha$ \And
  Yizhong Wang $^{\omega}$ \AND
  Hannaneh Hajishirzi $^{\alpha\omega}$ \And
  Matthew Peters $^\alpha$ \AND
  \textnormal{$^\alpha$Allen Institute for AI} \\
  $^\omega$Paul G.~Allen School of Computer Science \& Engineering, University of Washington \\
  \texttt{\{hamishi,akshitab,yizhongw,hanna,matthewp\}@allenai.org}}

\begin{document}
\maketitle
\begin{abstract}
Recent NLP models have shown the remarkable ability to effectively generalise `zero-shot' to new tasks using only natural language instructions as guidance.
However, many of these approaches suffer from high computational costs due to their reliance on concatenating lengthy instructions with every input example, resulting in costly reprocessing of the instruction.
To avoid this, we introduce Hypernetworks for INstruction Tuning (HINT), which convert task instructions and examples into parameter-efficient modules inserted into an underlying model using a pretrained text encoder, eliminating the need to include instructions in the model input.
The hypernetwork in HINT also produces an encoded instruction, which we concatenate with encoded inputs during decoding to further improve performance.
HINT models outperform strong state-of-the-art baselines by over 10\% when controlling for compute (measured in FLOPs).
By converting instructions into modules, HINT models can effectively disregard the length of instructions and few-shot example inputs in terms of compute usage. As a result, HINT can enhance its performance by up to 25\% by incorporating additional few-shot data, while utilizing only up to 5\% more compute. This combines the strengths of parameter-efficient fine-tuning and in-context learning. We release our code publicly\footnote{Our code is available at: \\ \href{https://github.com/allenai/hyper-task-descriptions}{\texttt{https://github.com/allenai/hyper-task-descriptions}}.}.
\end{abstract}
\section{Introduction}

\begin{figure}[h]
    \centering
    \adjustbox{max width=.5\textwidth}{
        \includegraphics{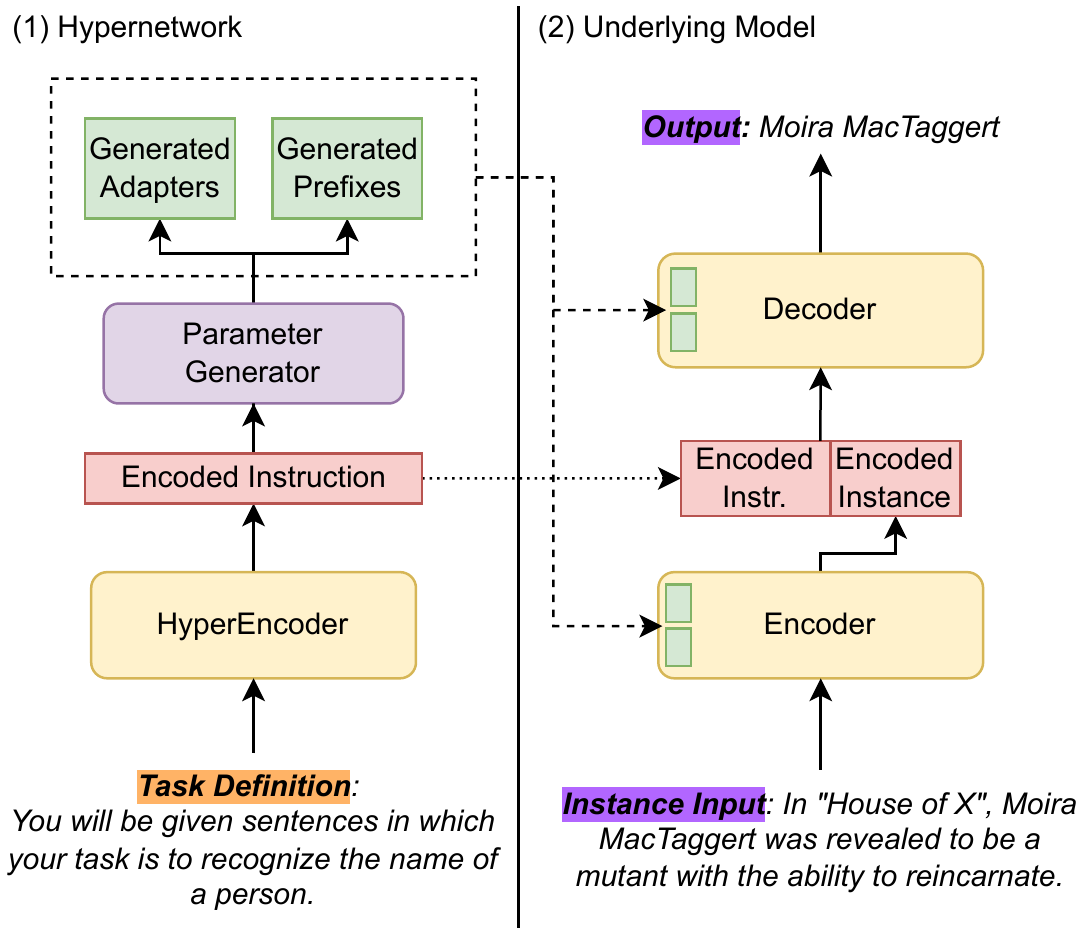}
    }
    \caption{Overview of HINT. (1) We feed an instruction into a HyperEncoder to produce an encoded instruction, use it to generate prefix and adapter weights, and then insert them into the underlying model. (2) We run the underlying model encoder as usual and optionally concatenate the encoded input with the previously encoded instruction, before running the underlying model decoder to generate the answer. We only use the hypernetwork once per task.}
    \label{fig:hint_overview}
\end{figure}

Large pretrained language models have demonstrated a striking ability to perform new tasks through the use of {\it in-context} examples or instructions alone \citep{gpt3}, or after training on input instances augmented with instructions \citep{weller-etal-2020-learning, 
 mishra-etal-2022-cross, sanh2022multitask, wei2022finetuned, flant5, supernaturalinstructions}. This ability allows a single model to adapt to many tasks where training data is difficult to collect or task-specific fine-tuning is impractical (i.e., `zero-shot' settings): models trained on instructions need only a single instruction to achieve non-trivial performance on the task at hand. The most common method to achieve this zero-shot ability is to meta-train the model with task instructions concatenated with every input, allowing the model to learn to associate instructions with tasks. While empirically highly successful, this is inefficient and  requires reprocessing lengthy task instructions and any additional task data (e.g., few-shot examples) with every input example. 

In this paper, we introduce Hypernetworks\footnote{Hypernetworks are neural networks trained to generate neural networks \citep{ha2017hypernetworks}.} for Instruction Tuning (HINT), which directly generate task-specific parameter-efficient modules given only an instruction, combining the benefits of instruction-based learning with parameter-efficient modules.
HINT models convert instructions and other task data (e.g., few-shot examples) into efficient modules within a pretrained language model, enabling cheaper inference and better compute scaling with few-shot data for an underlying instruction-based meta-learning approach. 
Additionally, fusing hypernetwork-encoded instructions with the encoded input at the underlying model decoder greatly improves the performance while using minimal extra compute.  An important benefit of HINT is that it processes instructions and other task information only once, making the compute used by our method almost independent of the amount of task data available, unlike both regular finetuning and input concatenation-based approaches (see Figure~\ref{fig:flops}).



We find that our hypernetwork-based approach (`HINT'), is able to achieve similar performance to baselines that receive the full instruction with every input example while using significantly less compute (as measured by FLOPs), due to the greatly reduced input length. When controlling for inference budget, we find that \textbf{HINT models outperform strong baselines in zero- and few-shot settings}. This validates our assumption that we can significantly reduce inference costs by avoiding reprocessing the instruction with every input, and instead saving it for repeated use. Furthermore, we find that including additional few-shot information alongside task instructions significantly improves HINT model performance while using minimal additional compute during inference. Ultimately, our work pushes towards directly generating cheap, customised models from task data, without requiring any expensive task-specific finetuning.

In summary, our findings are:
\begin{itemize}
    \item We introduce HINT models, which make use of a text-conditioned hypernetwork to generate parameter-efficient modules based on task descriptions and few-shot examples.
    \item HINT models, by reducing input lengths, are able to achieve similar performance to strong full-input baselines while reducing inference cost (measured in FLOPs) by up to 4$\times$.
    \item As the compute used by HINT models is effectively independent of the length of the instruction and amount of few-shot data provided with the instruction, HINT models provided with additional few-shot data simultaneously outperform and use up to 4$\times$ fewer FLOPs than baselines without few-shot data.
    \item HINT models outperform strong decoder-only baselines. While decoder-only models allow for input caching, we find that instruction-tuned GPT-2 models significantly underperform HINT models (8-9 point difference), matching prior work suggesting that encoder-decoder models work better for instruction-tuning \citep{obj, opt_iml}.
\end{itemize}

\section{Related Work}

\paragraph{Instruction Following} Further finetuning large pretrained language models on instructions has been found to greatly improve zero-shot generalisation, as the finetuned model learns to make use of the instructions to perform the given task \citep{weller-etal-2020-learning, wei2022finetuned, mishra-etal-2022-cross, flant5, supernaturalinstructions}. Additionally, \citet{sanh2022multitask} found that training models on multiple prompts per task also resulted in improved performance, suggesting that further increasing prompt diversity aids generalisation, even when using the same pool of underlying tasks.
The majority of these popular instruction-tuning approaches involve concatenating the instruction with the input directly and training a text-to-text model on these combined inputs. As the instruction can be as long as, if not longer, than the input\footnote{As is the case for Super-Natural Instructions, see Appendix~\ref{sec:median_calcs}.}, this can greatly increase the computation needed to process inputs compared to task-specific models.

\paragraph{In-Context Learning} Similar to instruction-based models, in-context learning \citep{gpt3}, where example instances are used in place of or in addition to instructions, also requires extremely long and expensive-to-process inputs for every test example, with \citet{liu2020tfew} showing that parameter-efficient finetuning (PEFT) can be cheaper and more effective when dealing with many test examples. In this work, we propose a halfway step between PEFT and instruction concatenation, where we train a model to predict parameter-efficient modules based on instructions, avoiding the few-shot training required by \citet{liu2020tfew} while also avoiding repeatedly processing lengthy inputs.

\paragraph{Hypernetworks In NLP} Hypernetworks \citep{ha2017hypernetworks, fwp} in NLP have recently gained popularity in multitask and multilingual setups due to their ability to softly share parameters while avoiding negative interference through the use of shared parameter generation module. Several approaches \citep{karimi-mahabadi-etal-2021-parameter, tay2021hypergrid, hyperprompt} learn per-task embeddings along with a shared hypernetwork to generate task-specific adapter or prefix modules. This means making the model perform new tasks requires at least few-shot learning to learn a task embedding. Recent work has explored using text-conditioned hypernetworks for parameter-efficient multitasking \citep{hyperdecoders} or improving out-of-domain generalisation \citep{volk2022example}, removing the need to train task-specific embeddings. Hypernetwork-based methods have also been highly successful in multilingual settings, where generating language-specific models via shared hypernetworks often results in improved performance across various tasks \citep[][\textit{inter alia}]{platanios-etal-2018-contextual, hyperadapters, Ustun2022HyperXAU}

Our work primarily builds on \citet{ye-ren-2021-learning}, which explored generating adapters from task descriptions. We expand their approach to larger models and datasets and find that pretraining and a significantly different hypernetwork architecture are important for achieving strong performance.

Our work is also similar to the concurrently developed \citet{hypertuning} and \citet{boostingmeta}, both of which examine how well hypernetwork-based meta-learning can improve model performance in zero- and few-shot settings. \citet{boostingmeta} examine hypernetworks and model-agnostic meta-learning for instruction-finetuning and find that they can yield improved performance on difficult unseen tasks in Super-Natural Instructions. However, they still struggle to achieve overall good zero-shot performance and do not investigate eliminating task descriptions from the model input itself. \citet{hypertuning} find that training a hypernetwork to produce model adaptations provides an initialisation better than pretraining for parameter-efficient adaptations and that this initialisation improves with more few-shot examples provided to the hypernetwork. They also explore eliminating the instruction from the underlying model input but find this severely underperforms baseline approaches. We have similar findings, but find that our novel hypernetwork design and use of instruction fusion closes the gap with baseline approaches. We also perform further analysis of the hypernetwork-based models and show that when controlling for inference compute budgets, our hypernetwork-based model still outperforms strong baselines.

\section{HINT Model}

Here, we introduce the main elements of our HINT model. The model has two core parts: the \textbf{hypernetwork}, which takes in text instructions and outputs parameter-efficient modules, and the \textbf{underlying model}, into which we insert the generated parameter weights. The underlying model is simply an encoder-decoder\footnote{We use encoder-decoder models as they generally outperform decoder-only models for zero-shot generalisation -- see Section~\ref{sec:sni_res}.} transformer model \citep{transformer} with additional parameter-efficient adaptations inserted in, while the hypernetwork has a more complex architecture which we describe below. Figure~\ref{fig:hint_overview} provides a visual overview.

\subsection{Hypernetwork}

The first step in our model is to make use of a hypernetwork to convert an instruction to parameter-efficient modules.  Our hypernetwork consists of three core elements: an \textbf{encoder} (or `\textbf{hyperencoder}') to transform instruction and few-shot text into continuous (contextual) representations, saving the encoded instructions for \textbf{instruction fusion} during decoding, and a \textbf{parameter generator} to then convert these embeddings into the parameter efficient modules.

\paragraph{HyperEncoder} To encode our text, we use a pretrained language model encoder. We initially experimented with using different encoder configurations, and find that re-using the encoder from the underlying model we wish to augment works well, and tying the hypernetwork and underlying encoder model weights works best.

\paragraph{Instruction Fusion} We save the instruction representations produced by the hyperencoder and allow the decoder of the underlying model access to them by concatenating them with input examples during inference and training. This is inspired by the fusion-in-decoder method used in open-domain QA \citep{izacard-grave-2021-leveraging}.

\subsubsection{Efficient Parameter Generators} 

\paragraph{Parameter Generators} Our generator design consists of two parts. First, we use a trainable set of embeddings and perform multi-head cross-attention with the encoded instruction and these embeddings. Each embedding represents a unique column or token in each parameter-efficient module (e.g., prefix, adapter - see below) for each layer. This allows us to effectively collect the information required for different parameters in different embeddings via cross-attention with the instruction:
\begin{align*}
    \text{embed} &= [\bm{\alpha}_{e^1_1}, \bm{\alpha}_{e^1_2}, ..., \bm{\alpha}_{e^2_1}, \bm{\alpha}_{e^2_2}, ..., \bm{\pi}_{e^1_1}, ...] \\
    \text{embed}' &= \text{Cross-Attention}(\text{embed}, \text{instr.})
\end{align*}
Where $\bm{\alpha}_{e^1_1}$ refers to an embedding we will use as the first column of the first layer adapter weight, $\bm{\alpha}_{e^1_2}$ is the second column, $\bm{\alpha}_{e^2_1}$ is the first column for the second layer adapter, $\bm{\pi}_{e^1_1}$ is the first token of the first layer prefix, etc.  We then take the subset of the embedding representing all columns/tokens for a particular model adaptation and pass it through a two-layer MLP to generate parameters. We use a unique network for each adaptation and share between layers (i.e. one network for prefixes for all layers, one for all adapter weights for all layers, etc.).
\begin{align*}
    \text{Adapter}_1 &= \text{reshape}[\text{MLP}_a(\bm{\alpha}'_{e^1_1}); \text{MLP}_a(\bm{\alpha}'_{e^1_2}); ...] \\
    \text{Prefix}_1 &= \text{reshape}[\text{MLP}_p(\bm{\pi}'_{e^1_1}); \text{MLP}_p(\bm{\pi}'_{e^1_2}); ...]
\end{align*}

Where Adapter$_1$ and Prefix$_1$ are the first layer adapter and prefix, respectively.

\paragraph{Generated Parameters} \label{sec:gen_params}We generate two types of parameter-efficient modules: \textbf{adapters}  \citep{pmlr-v97-houlsby19a} and \textbf{prefixes} \citep{li-liang-2021-prefix}. \textbf{Adapters} \citep{pmlr-v97-houlsby19a} are small bottleneck networks inserted into a transformer model. We follow \citet{he2022towards} in placing our adapters parallel to the feed-forward layer:
\begin{align}
    \bm{x} = \text{FFN}(\bm{x}) + f_1(\text{GELU}(f_2(\bm{x})))
\end{align}
Where $f_1, f_2$ are linear layers that project an input $\bm{x}$ to a small bottleneck size $n_a$ and then back up to the hidden size of the model respectively. \textbf{Prefixes} \citep{li-liang-2021-prefix} are short continuous sequences concatenated with the key and values in the self- and cross-attention modules in every layer of the underlying model.

\paragraph{Scaling Down Parameters} A na\"ive hypernetwork implementation may suffer from poor scaling with the size of the parameter-efficient modules. Consider a case where we wish to convert a single embedding of size $n_e$ to an adapter weight matrix of size $n_d \times n_a$ (the model hidden dimension size by adapter bottleneck size). Our hypernetwork generator will have $n_d*n_e*n_a$ parameters, and so increasing the adapter bottleneck $n_a$ quickly becomes extremely expensive, especially if $n_d$ is large - as is the case for large language models. We address this by decomposing the adapter weight into columns and \textit{assigning an embedding per column}. Thus, our hypernetwork now has to convert a sequence of embeddings with size $n_a \times n_e$ to an adapter weight of size $n_a \times n_d$, meaning that the network only needs $n_e*n_d$ parameters. This means that the size of our parameter-efficient modules is independent of the size of the hypernetwork, and we can effectively scale the size of our adapters or number of prefixes \textit{without extreme parameter blowup}. Note that we set $n_e = n_d$ in our experiments for simplicity.

\subsection{Underlying Model}

Once our hypernetwork has produced a set of parameter-efficient modules, we then insert these into our underlying network, and can then perform training and inference as normal. The underlying model can be any pretrained encoder-decoder model that works with our parameter-efficient modules. We make use of T5 \citep{t5} as our underlying model in our experiments.

\subsection{Training and Inference}

\paragraph{Hypernetwork Pretraining}
\label{sec:pretraining_exp}

\begin{figure}
    \centering
    \adjustbox{max width=.45\textwidth}{
        \includegraphics{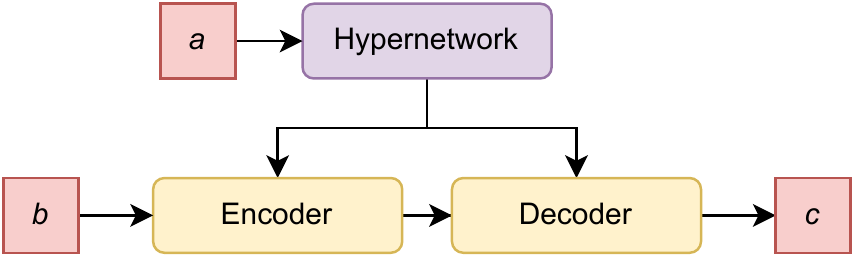}}
    \caption{Overview of our proposed pretraining scheme.}
    \label{fig:pretraining}
\end{figure}

To help better generalization, we pretrain the hypernetwork on a large corpus \citep[C4;][]{t5} before finetuning on multitask prompted datasets. Given a single input string, we split our input string into random-length chunks $a$, $b$, $c$, and feed $a$ to the hypernetwork, $b$ to the main network, and predict $c$. This resembles the input in used in instruction finetuning (as the instruction precedes the input in the default prompt used for Super-Natural Instructions). We fully finetune all parameters during pretraining.

\paragraph{Training HINT} HINT training looks similar to pretraining, except we replace $a$ with the task instruction (and any few-shot examples), $b$ with the main input, and $c$ with the gold generation. We used mixed-task batches such that a unique adaptor and prefix set is generated for every input in each batch.

This means, for every batch, we first generate a set of adapters, prefixes, and encoded instructions from a batch of tasks using the hypernetwork. The adapters and prefixes are placed within the underlying model to act as parameter-efficient modules (i.e., insert them into the model), and the encoded instruction is concatenated with encoded inputs during decoding. We then perform a forward pass of the underlying model with the inputs associated with each task in the batch and perform backpropagation using cross-entropy loss as standard for text-to-text models. As we fully finetune all parameters, the parameter generator will produce different weights for the same task inputs after a gradient step, meaning that we have to rerun the hypernetwork for every batch. This means that HINT requires more compute to train than a baseline transformer - although it provides significant compute reductions during inference, as we will see.


\paragraph{Inference} The inference process is similar to training, but we do not use mixed-batch inputs: instead, we generate the parameters for one task, insert them into the underlying model, and then process all test-time inputs for that task. This prevents redundant processing of the instruction.

We also consider the cost of HINT models during inference. We consider a case where we have to process $n$ samples from a single task. Assume each sample has length $i$ and the task instruction has length $t$. We will ignore the cost of processing (typically short) output sequences. Following prior work \citep{scaling_laws, liu2020tfew}, we use FLOPs as an estimate of the amount of compute required to run particular models and estimate that processing a token with an encoder-decoder model takes $N$ FLOPs to process a single token, where $N$ is the total number of model parameters.

In this scenario, a standard instruction-trained model which concatenates every input with the instruction (e.g., Tk-Instruct) uses $Nn(t+i)$ FLOPs to process all examples. Meanwhile, HINT models process the task instruction only once and so use roughly $N(t+ni)$ FLOPs\footnote{When reporting FLOPs, we use a more detailed formula described in Appendix~\ref{sec:compute_costs} that takes into account extra (albeit small) hypernetwork costs.}. This makes clear that HINT models (a) \textbf{scale better with more same-task inference examples} than input concatenation approaches (increasing $n$), and (b) \textbf{require relatively few extra FLOPs to process long instructions} (large $t$), allowing them to benefit from adding more few-shot examples without incurring significant compute increases.

\section{Experimental Details}
\label{sec:exp_details}

We evaluate our approach on two popular instruction-based datasets: \textbf{Super-Natural Instructions (SNI)} \citep{supernaturalinstructions} and the \textbf{T0 split of P3} \citep{sanh2022multitask, bach2022promptsource}. We use \texttt{t5x} and \texttt{seqio} \citep{t5x} to handle data preprocessing and model training. We use T5 v1.1 + LM adaptation \citep{lester-etal-2021-power} as our base models, using the 3B size unless otherwise stated. Unless otherwise stated, the hypernetwork generates prefixes of length 30 and adapters with a bottleneck size of 512, matching the sizes recommended by \citet{he2022towards}.  We use the Adafactor optimizer \citep{adafactor} with a constant learning rate of 0.001. Unless otherwise stated, we report results from single runs.

\paragraph{Pretraining} We pretrain all models for 10,000 steps (7,000 for 11B size models) using C4 \citep{t5} with a batch size of 1,024 samples and sequences of length 512. 

\paragraph{Super-Natural Instructions (SNI)} For SNI, we examine two settings: providing the hypernetwork with the task definition and the underlying network with the instance input only (\textbf{`Def'}), and providing the hypernetwork with the task definition and two few-shot task examples (\textbf{`Def + 2 Pos.'}). To train, we finetune our pretrained HINT models for 1,000 steps with a batch size of 1,024, with a maximum sequence length of 1,024 for both the underlying model and the hypernetwork input. We then evaluate the final checkpoint on the test split of SNI, which is a set of 119 unseen tasks. We use v2.6 of Super-Natural Instructions.

\begin{figure*}[]
    \centering
    \adjustbox{max width=\textwidth}{
    \includegraphics{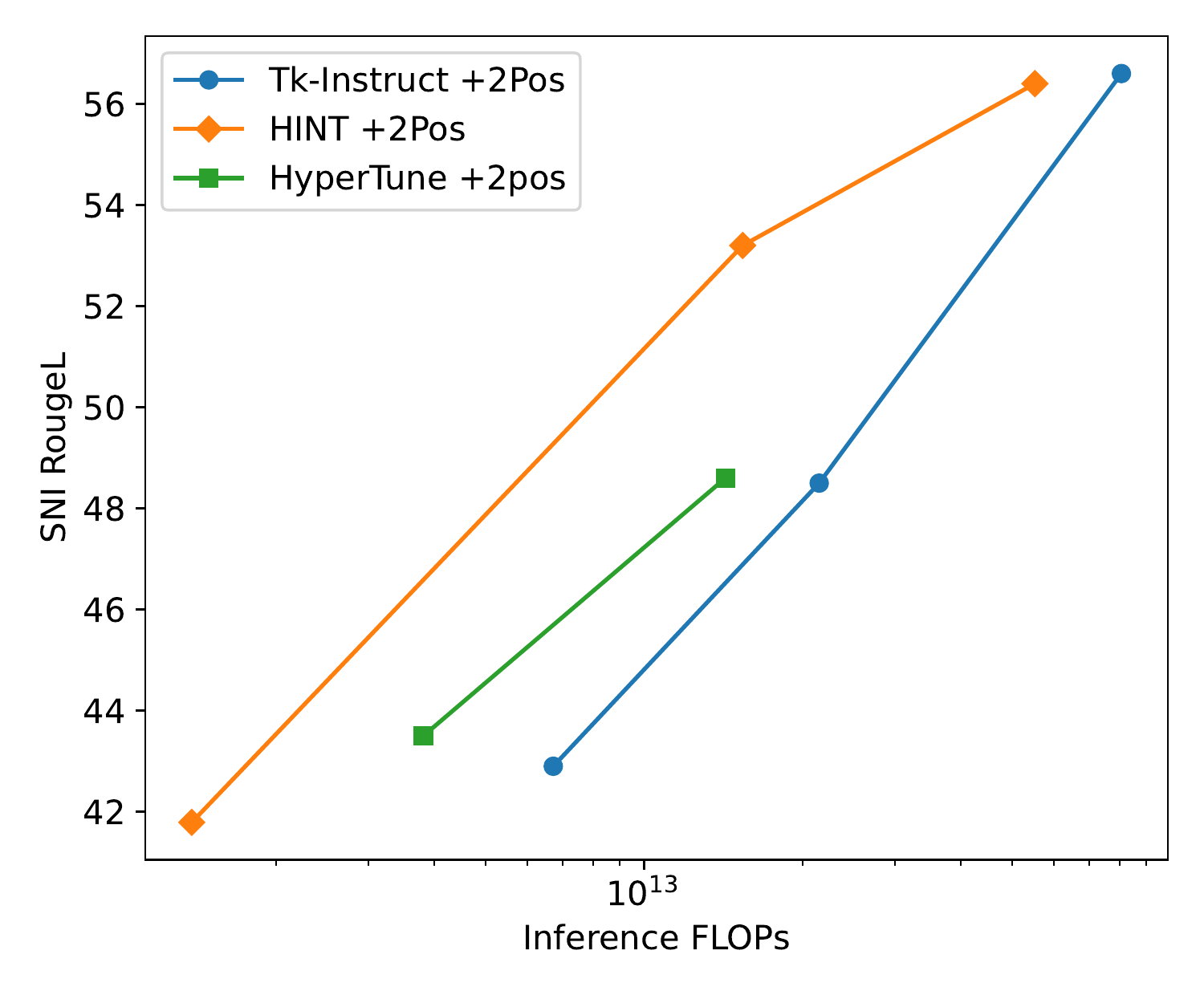}
    \includegraphics{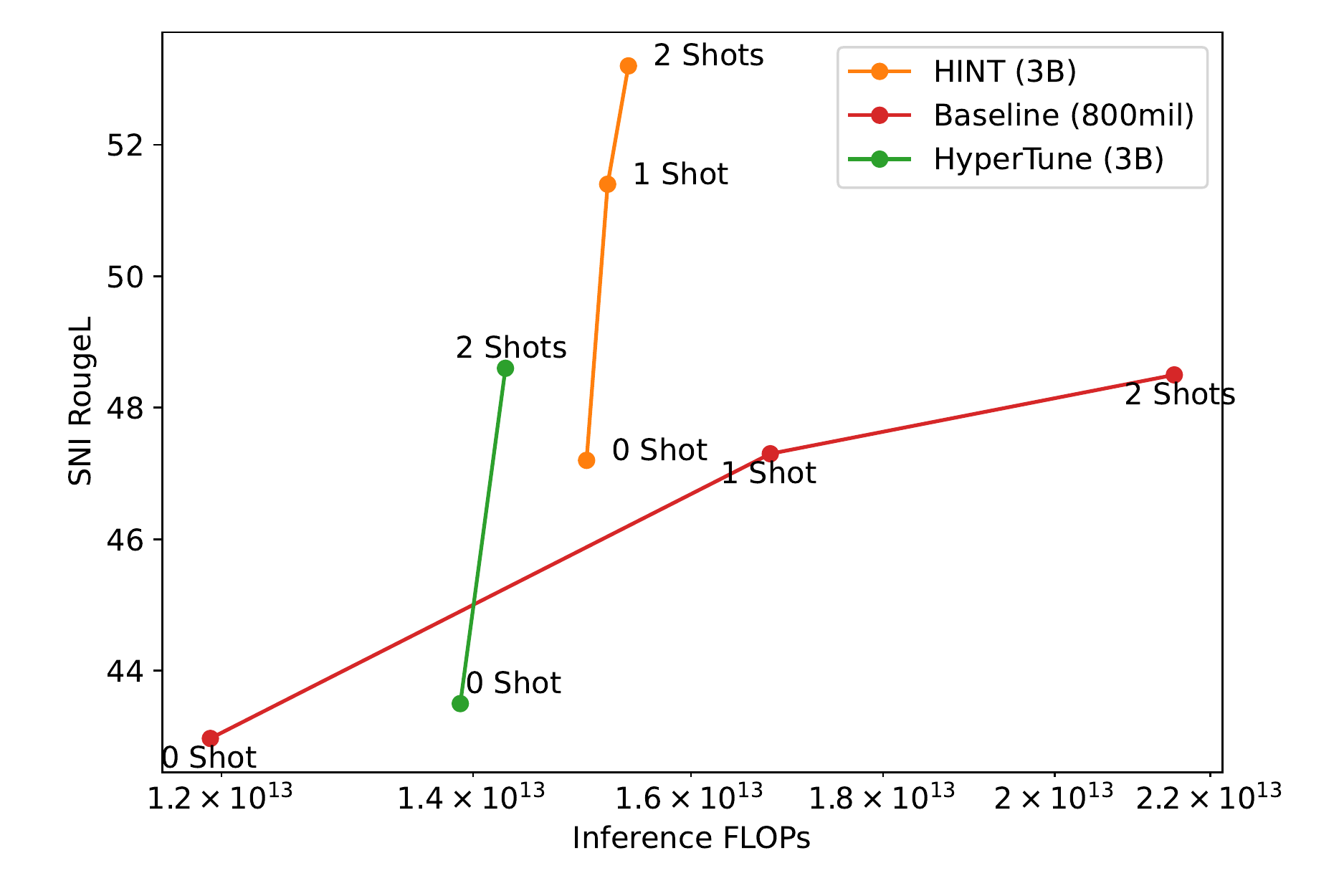}
    \includegraphics{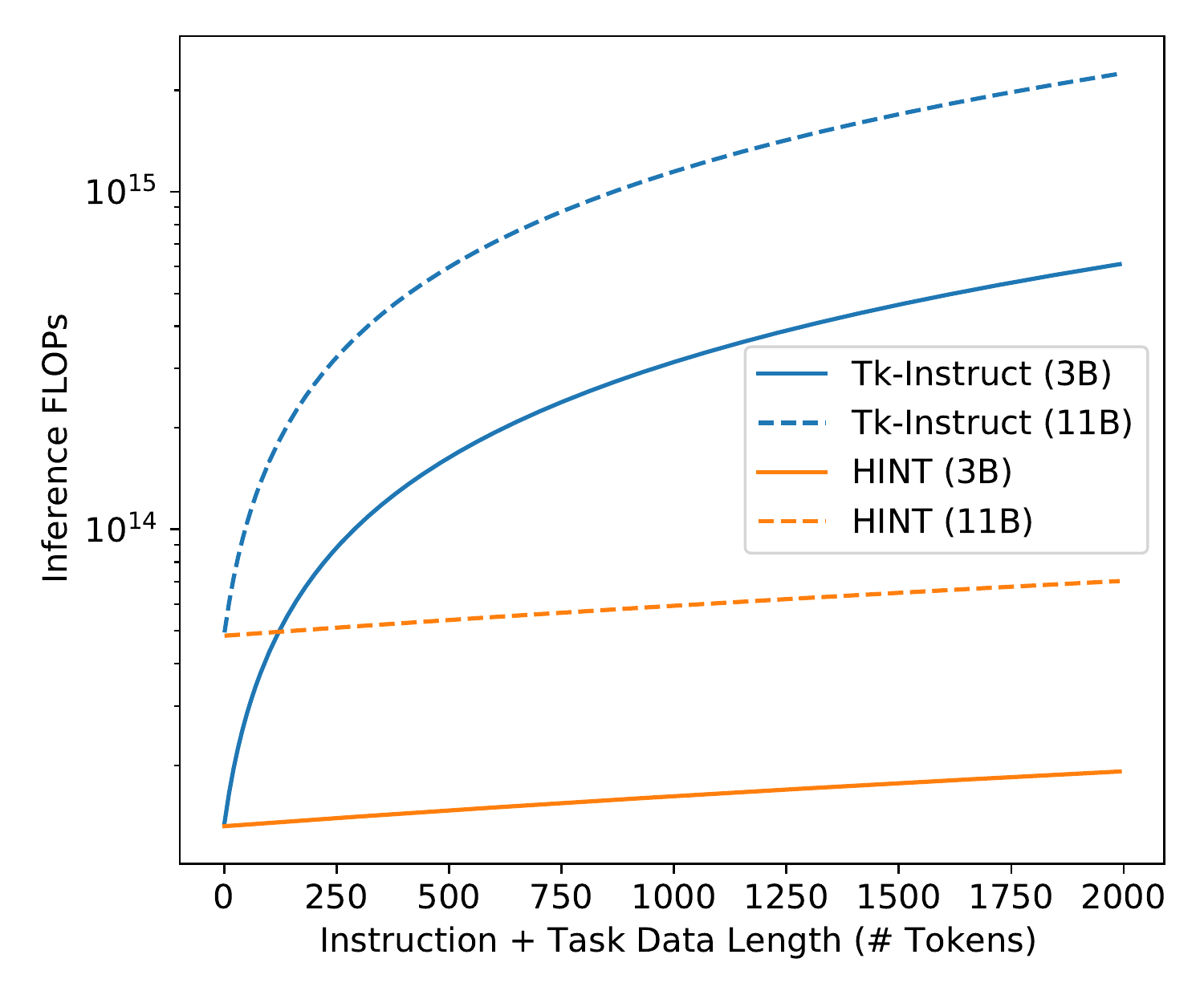}}
    \caption{(Left) SNI RougeL against FLOPs for varying model sizes using task definitions with two positive examples; (Centre) SNI RougeL against FLOPs for differing numbers of additional few-shot information; (Right) FLOPs against instruction and task data length (in number of tokens). FLOPs calculations are based on processing 100 examples from the same task during inference.}
    \label{fig:flops}
\end{figure*}

\paragraph{P3} For P3, we explore two settings: (a) \textbf{`joint'}, where we give the hypernetwork a templated form of the prompt with instance information removed and give the underlying model the full prompted input. (b) \textbf{`split'}, where we give the hypernetwork the templated prompt without instance information and give the underlying model only the instance information \textit{without} the prompt. In both cases, we fully finetune our model for 10,000 steps with a batch size of 2,048. We use a maximum input length of 1,024 for the underlying model and 512 for the hypernetwork. We train and evaluate on the same tasks and splits as T0 \citep{sanh2022multitask}.

\paragraph{Baselines} We primarily compare against \textbf{T0} and \textbf{Tk-Instruct}, models fully-finetuned on P3 and SNI respectively with all task information concatenated with the input. We replicate these models, matching the finetuning settings used for HINT models, and find that our replications significantly outperform previously reported results, making these baselines extremely strong. We note where results are our replications or reported from prior work. We additionally compare against `\textbf{X + PEFT}', the prior models with adapters and prefixes added in before finetuning, \textbf{HyperTune} \citep{hypertuning}, a concurrent work that primarily makes use of a pretrained hypernetwork but without instruction fusion, \textbf{Hypter} \citep{ye-ren-2021-learning}, a prior hypernetwork-based model that does not use pretraining, \textbf{GPT-2} \citep{radford2019language}, a strong decoder-only model, which we fully finetune, and \textbf{OPT} \citep{zhang2022opt}, another strong decoder-only model, which we also fully finetune.

\section{Model Performance and Efficiency}
\label{sec:results}
\begin{table*}[t]
\centering
\adjustbox{max width=.8\textwidth}{
\begin{tabular}{@{}lccccc@{}}
\toprule
 & & \multicolumn{2}{c}{\textbf{Def}} & \multicolumn{2}{c}{\textbf{Def + 2 Pos.}}\\
\cmidrule(r){3-4} \cmidrule(r){5-6}\textbf{Model}&\textbf{Model Size} & \textbf{RougeL} & \textbf{Rel. FLOPs} & \textbf{RougeL} & \textbf{Rel. FLOPs} \\\midrule
Tk-Instruct (our replication) & 250 mil. & \textbf{35.3} & $\times$1.0 & \textbf{42.9} & $\times$1.5 \\
Tk-Instruct \citep{supernaturalinstructions} & 250 mil. & - & - & 42.1 & $\times$1.5 \\
Tk-Instruct + PEFT & 250 mil. & 33.3 & $\times$1.1 & \textbf{42.9} & $\times$1.6 \\
Hypter (our replication) & 250 mil. & 12.1 & $\times$0.4 & 10.6 & $\times$0.4 \\
\textbf{HINT (ours)} & 250 mil. & 33.3 & \textbf{$\times$0.4} & 41.8 & \textbf{$\times$0.4}  \\ \hdashline
Tk-Instruct (our replication) & 3B & 48.9 & $\times$12.0 & \textbf{56.6} & $\times$17.9 \\
Tk-Instruct \citep{supernaturalinstructions} & 3B & 45.0 & $\times$12.0 & 54.3 & $\times$17.9 \\
Tk-Instruct + PEFT & 3B & \textbf{49.8} & $\times$12.4 & 56.2 & $\times$18.5 \\
No-Instruct & 3B & 12.4 & \textbf{$\times$3.9} & - & - \\
GPT-2 XL & 1.5B & 38.2 & $\times$4.1 & 45.3 & \textbf{$\times$4.2} \\
Hypter (our replication) & 3B & 16.8 & $\times$4.3 & 14.2 & $\times$4.4 \\
HyperTune \citep{hypertuning} & 3B & 38.9 & $\times$4.1 & 48.6 & $\times$4.3 \\
\textbf{HINT (ours)} & 3B & 47.2 & $\times$4.5 & 53.2 & $\times$4.6  \\ \hdashline
Tk-Instruct (our replication) & 11B & 53.6 & $\times$44.0 & 60.5 & $\times$65.7 \\
Tk-Instruct \citep{supernaturalinstructions} & 11B & - & - & \textbf{62.0} & $\times$65.7 \\
Tk-Instruct + PEFT & 11B & \textbf{54.6} & $\times$44.0 & 60.3 & $\times$65.7 \\
OPT-13B & 13B & 44.8 &  $\times$15.9 & 51.5 & \textbf{$\times$16.4}\\
Hypter (our replication) & 11B & 15.5 & \textbf{$\times$15.3} & 13.4 & $\times$15.7 \\
\textbf{HINT (ours)} & 11B & 51.1 & $\times$16.1 & 56.4 & $\times$16.5  \\
\bottomrule
\end{tabular}}
\caption{Super-Natural Instructions RougeL and relative number of FLOPs used when given task definition only (`Def'), and task definition along with 2 labelled examples (`Def + 2 Pos.'). Where noted, results are taken directly from \citet{hypertuning} and \citet{supernaturalinstructions}. Relative FLOPs cost is calculated relative to the base-size Tk-Instruct with task definition only. We calculate the values using the number of FLOPs required to process 1 task with 100 examples for each model. Model size is given by the number of parameters.}
\label{tab:sni_res}
\end{table*}

\subsection{Super-Natural Instructions}
\label{sec:sni_res}
We report the performance and inference costs of HINT models and baselines in Table~\ref{tab:sni_res} and Figure~\ref{fig:flops} and find that:

\textbf{HINT models outperform baselines when FLOPs-matched.} As seen in Figure~\ref{fig:flops} (left), when FLOPs-matched, HINT models outperform Tk-Instruct, a strong baseline that fully concatenates the instruction with every input. This holds for both `Def' and `Def + 2 Pos.' settings.

\textbf{HINT models are up to 4$\times$ more efficient than similarly-sized baselines.} we find that HINT models use 2--4x fewer FLOPs than similarly-sized state-of-the-art Tk-Instruct baselines (Table~\ref{tab:sni_res}). While other hypernetwork-based models are able to achieve similar compute savings, their performance is significantly worse than HINT ($\geq 8$ points). HINT has similar cost to a model trained without including instructions in the input (`No-Instruct'), while performing over 30 points better.

\textbf{HINT models improve performance with few-shot examples, but do not cost more FLOPs.} When introducing additional few-shot data (`Def + 2 Pos.'), HINT models improve dramatically (5-8 points) but the compute used barely increases (Figure~\ref{fig:flops}, centre), as HINT models only need to encode the task data (instruction and few-shot examples) once per task. In contrast, while Tk-Instruct similarly improves with few-shot examples, the compute needed during inference increases dramatically, usually costing around $1.5\times$ more. Overall, we find that HINT models require much less compute to deal with longer instruction and few-shot data inputs than Tk-Instruct (Figure~\ref{fig:flops}, right).

\textbf{HINT models outperform a strong decoder-only baseline.} HINT significantly outperforms GPT-2 and OPT-13B, in line with prior work that shows encoder-decoder models often significantly outperform even much larger decoder-only equivalents \citep{obj, opt_iml}. In particular, 11B-size HINT outperforms OPT-13B by 5 points or more despite using a similar number of FLOPs. This highlights the utility of improving efficiency for encoder-decoder-based models. We also note that caching key/value attention pairs, the simplest way to reduce inference costs with decoder-only models, scales worse than HINT. The size of cached key/value pairs for GPT-2 is $\propto lds$, where $l$ is the number of layers, $d$ is the size of the model hidden dimension, and $s$ is the cached sequence length. In contrast, the size of the saved PEFT parameters for HINT is  $\propto sd + ld$, which scales better with respect to sequence length (larger $s$) and model size (larger $d$, $l$)\footnote{We provide details for these calculations in Appendix \ref{sec:memory_costs}.}.

\begin{table}[h]
\centering
\adjustbox{max width=.4\textwidth}{
\begin{tabular}{lcc}
\toprule
 \textbf{Model} & \textbf{Avg} & \textbf{Rel. FLOPs} \\\midrule
T0-3B & 54.9 & $\times$1.0  \\
T0-3B (our replication) & 64.4 & $\times$1.0 \\ 
T0-3B + PEFT & \textbf{65.5} & $\times$1.0 \\ 
No Prompt & 57.5 & \textbf{$\times$0.8} \\ \hdashline
Hypter (Joint) & 64.6 & $\times$1.0 \\
\textbf{HINT (Joint)} & 65.4 & $\times$1.1 \\
Hypter (Split) & 56.2 & \textbf{$\times$0.8}  \\
\textbf{HINT (Split)} & 60.3 & \textbf{$\times$0.8} \\
\bottomrule
\end{tabular}}
\caption{Avg performance over T0 evaluation tasks after training on the T0 P3 train set. FLOPs are calculated assuming we are processing 100 examples of a single task. The `Joint' and `Split' HINT variants refer to the two input formats for P3 described in Section~\ref{sec:exp_details}.}
\label{tab:p3_res}
\end{table}

\begin{table}[h]
\centering
\adjustbox{max width=.48\textwidth}{
\begin{tabular}{cccccc}
\toprule
\textbf{\# Shots} & \textbf{T0-3B (our repl.)} & \textbf{HINT (split)} & \thead{\bf HINT FLOPs \\ \bf Reduction} \\\midrule
1 & \textbf{66.4} & \textbf{66.4} & $\times$2.3 \\
2 & \textbf{67.1} & 66.6 & $\times$3.2 \\
4 & 67.1 & \textbf{67.2} & $\times$5.1 \\
5 & \textbf{67.9} & 67.1 & $\times$6.0 \\
\bottomrule
\end{tabular}}
\caption{P3 performance with differing numbers of few-shot examples using 3B size models and the FLOPs reduction when using HINT instead of T0-3B for that number of shots. In few-shot settings, HINT always remains wihin 1 point of T0-3B despite the greatly reduced FLOPs cost.}
\label{tab:p3_few_shot}
\end{table}

\subsection{P3}
We report results on the T0 evaluation set in Table~\ref{tab:p3_res}, with full results in Appendix~\ref{app:p3_full}. We find that:

\textbf{Our T0-3B replication significantly outperforms the results reported by \citet{sanh2022multitask}.} This matches prior suggestions that T0 is undertrained \citep{hypertuning, wu-etal-2022-continued}. We provide further details in Appendix~\ref{sec:t0_rep}.

\textbf{HINT outperforms hypernetwork baselines.} The HINT model consistently outperforms Hypter, a prior hypernetwork-based approach, and learns to make use of the P3 prompts as evidenced by its improved performance over a baseline model trained without prompts (`No Prompt'). 

\textbf{HINT remains cheaper than T0 for inference.} HINT uses significantly less flops than T0-3B, albeit with smaller savings compared to SNI, likely due to the different style of prompts: P3 prompts tend to be shorter, and interleave task inputs (e.g. `Does <sentence 1> entail <sentence 2>?'). Despite this, HINT still provides reasonable FLOPs savings. We suggest that the performance of HINT could be greatly improved by leveraging additional few-shot information, further exploiting the efficiency of HINT models in encoding task data.

We investigate if HINT models can provide benefits even when the input and instruction are concatenated through training and evaluating in the `joint' setting of P3, and find that HINT performs similarly to T0-3B with additional parameter-efficient modules, which suggests that the hypernetwork is unable to improve on the baseline model through additional customisation, and so is primarily useful as a mechanism for reducing inference costs and cheaply incorporating few-shot data.

\textbf{HINT performs similarly to the baseline in few-shot settings.} In Table~\ref{tab:p3_few_shot}, we show that HINT remains within 1 point performance of T0 in few-shot settings, despite the large reductions in FLOPs cost, using up to 6$\times$ fewer FLOPs. This makes HINT especially useful in few-shot scenarios.

\section{Analysis}

\subsection{Pretraining}

\begin{table}[]
\centering
\adjustbox{max width=.4\textwidth}{
\begin{tabular}{@{}lcc@{}}
\toprule
 \textbf{Model} & \textbf{Pretraining} & \textbf{SNI RougeL}\\ \midrule
HINT & None & 44.0 \\
HINT & Ours & \textbf{46.3} \\
HINT & CACLM & 45.8 \\ \hdashline
HINT - No Instr. Fus. & None & 27.4 \\
HINT - No Instr. Fus. & Ours & \textbf{32.1} \\
HINT - No Instr. Fus. & CACLM & 30.4 \\ \bottomrule
\end{tabular}}
\caption{SNI performance for HINT models with and without instruction fusion after 10,000 steps of the given pretraining scheme and 1,000 steps of finetuning on SNI. CACLM is the pretraining scheme proposed by \citet{hypertuning}.}
\label{tab:pretraining}
\end{table}

We compare using no pretraining, our pretraining scheme, and the pretraining scheme proposed by \citet{hypertuning} (`CACLM') in Table~\ref{tab:pretraining}. As the pretraining scheme is primarily for improving the parameter generators, we evaluate its effect both with and without using instruction fusion (`HINT' and `HINT - No Instr. Fus.', respectively). 

We find that: (a) \textbf{using pretraining gives a large boost in performance for hypernetwork-only models}, showing that pretraining is essential to good hypernetwork performance, and (b) \textbf{using our pretraining scheme works best overall}. We hypothesise this reflects the fact that our scheme is closer to the Super-Natural Instructions format than CACLM. Unlike \citet{hypertuning}, we found that further pretraining did not aid performance. This is likely due to the fact that we tie the underlying model encoder and hypernetwork encoder weights together, meaning that the model weights must balance between acting as the hypernetwork and underlying model encoder.

\subsection{Inference Speed}

\begin{figure}
    \centering
    \adjustbox{max width=\linewidth}{
    \includegraphics{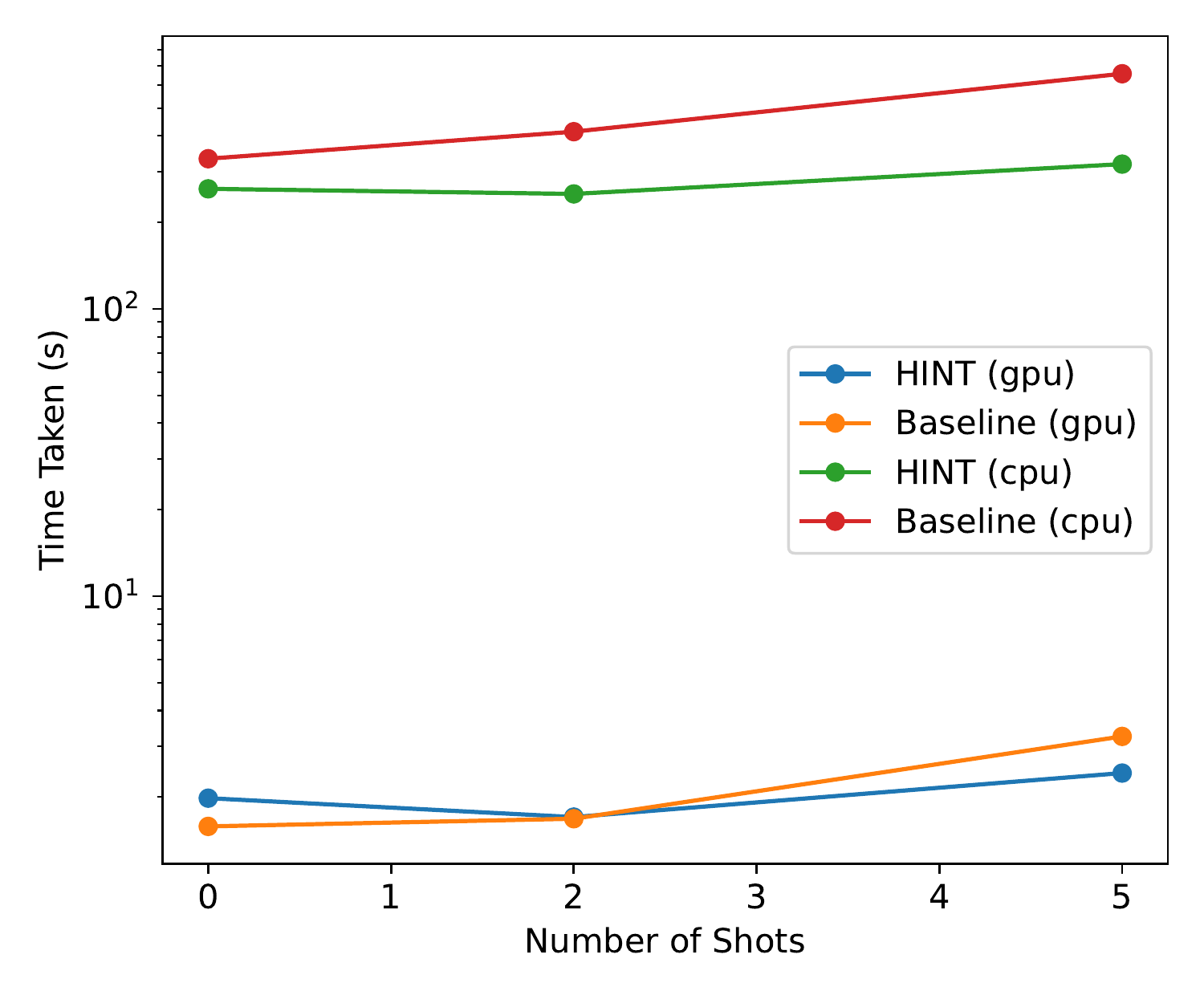}
    }
    \caption{Time taken to process 100 examples with average SNI lengths with varying numbers of shots on CPU and GPU for 3B HINT and baseline (i.e., vanilla T5) models.}
    \label{fig:shots_timing}
\end{figure}

While HINT provides significant FLOPs reductions compared to baselines, these do not necessarily translate to real-world inference speedups. We examine this by measuring the average speed of HINT to process 100 samples of the same task, assuming the average input lengths given in Appendix~\ref{sec:median_calcs}.

As seen in Figure~\ref{fig:shots_timing}, while baseline decoding remains faster for small input lengths on GPU\footnote{This is likely due to the small additional overhead of running the HyperEncoder, which must be run before the rest of the model.}, it lags compared to HINT for longer sequences. In fact, HINT's inference latency increases at a much slower rate compared to the baseline as the input size increases (with the number of shots), highlighting that HINT is especially effective in few-shot scenarios and scenarios with lengthy inputs.

\subsection{Architecture Ablations}

\begin{table}[]
\centering
\adjustbox{max width=.32\textwidth}{
\begin{tabular}{@{}lcc@{}}
\toprule
 \textbf{Model} & \textbf{SNI RougeL} \\ \midrule
 Adapters + Prefixes & \textbf{32.1} \\
 Adapters ($a = 512$) & 30.1  \\
 Prefixes ($l = 30$) & 12.1 \\
 Prefixes ($l = 512$) & 15.1 \\
 LoRA ($r = 128$) & 12.1 \\
 LoRA ($r = 512$) & 12.6 \\ \bottomrule
\end{tabular}}
\caption{SNI performance of different parameter-efficient modules in a HINT model without instruction fusion. $a$, $l$, $r$ are the bottleneck size, number of tokens, and rank used for each experiment respectively.}
\label{tab:ablation}
\end{table}

We experiment with a series of ablations to determine the best architecture for HINT, and find that:

\paragraph{Adapters and prefixes work best together.} We consider alternatives to using adapters and prefixes together: using adapters alone, using prefixes alone, and using LoRA \citep{hu2022lora} instead of either. In order to isolate the effect of these choices, we test without using instruction fusion. We find that adapters and prefixes provide the best overall performance, with prefixes-only and LoRA-only performance substantially worse, even when increasing the number of parameters generated. This suggests that our hypernetwork approach is more adept at generating certain types of PEFT modules.

\begin{table}[]
\centering
\adjustbox{max width=.27\textwidth}{
\begin{tabular}{@{}lcc@{}}
\toprule
 \textbf{Model} & \textbf{SNI RougeL} \\ \midrule
 HINT & \textbf{47.2} \\
 + Decoder & 42.6 \\
 - Instr. Fus. & 32.1 \\
 - PEFT Gen. & 40.9 \\ \bottomrule
\end{tabular}}
\caption{HINT model ablations. All models are pretrained for 10,000 steps, except for - PEFT Gen., which contains no new parameters and requires no pretraining.}
\label{tab:ablation}
\end{table}

\paragraph{PEFT and instruction fusion are complementary.} We find that using just the generated parameter-efficient modules or the encoded instruction alone (`-Instr. Fus.' and `-PEFT Gen.' in Table~\ref{tab:ablation}) perform significantly worse than using both methods together, suggesting that these methods provide complementary improvements.

\paragraph{Cross-Attention Layer wins over Full Decoder.} We compare using a full T5 decoder (with self-attention removed) as the hypernetwork weight generator as in \citet{hypertuning} with our approach, and find that our single multi-head cross-attention layer performs better at a much cheaper cost than using the full decoder (`+ Decoder' in Table~\ref{tab:ablation}).

\section{Conclusion}

We introduce Hypernetworks for INstruction Tuning (HINT) models and show that they consistently outperform strong full-input baselines when controlling for inference compute. This is primarily due to the fact that HINT models process their task instructions once per task, while current state-of-the-art models re-encode instructions with every task input. We show that the success of HINT models relies on a pretrained hypernetwork, which converts task instructions into parameter-efficient modules and an encoded instruction, both of which we insert into the underlying model.

Future work could investigate how HINT aids in few-shot settings, further building on HINT's strong few-shot efficiency and taking advantage of the improved initialisation provided by hypernetworks \citep{hypertuning}. Overall, HINT models combine the benefits of parameter-efficient learning with the benefits of instruction-based learning, allowing one to easily turn pretrained language models into efficient, task-customised models.

\section*{Limitations}
While promising, HINT comes with several drawbacks related to its ease of use. First, HINT takes advantage of the fact that (a) instructions are often long, and (b) often we want to perform inference over a larger ($> 100$) amount of examples with the same instruction. If either of these items are not true in a setup, then HINT is unlikely to provide a large benefit over simply including the instruction with the input text. This can be seen in the smaller compute savings provided by HINT for P3 in Table~\ref{tab:p3_res}. Second, while HINT is compute-efficient at inference time, it is far more costly to train, as it effectively requires running the underlying model together with the hypernetwork for every batch. This means that while HINT may be useful for practitioners with limited compute budgets, it may be difficult to train HINT models with the same limited budget. Finally, we train and test on English data only, and do not explore the generalisation of our approach to multilingual setups. Considering the success of hypernetworks in multilingual settings \citep{platanios-etal-2018-contextual, hyperadapters, Ustun2022HyperXAU}, we believe this is a promising direction for future research. As such, while promising, HINT is limited by certain assumptions made about the length and format of instruction-augmented data, and we hope further improvements of the method work towards loosening these assumptions.

\section*{Ethics Statement}
We believe that the impact of our work is largely beneficial, examining a novel method to make instruction-based models cheaper to use. This may aid in reducing the carbon footprint of large language models running in inference \citep{greenai} and in making these models more accessible to people with limited compute budgets. However, we also note that our approach requires unsupervised pretraining on a large corpus, making it difficult to document exactly the data it has seen during training and making it likely to reflect problematic or even dangerous biases within the corpus \citep{parrots}. We believe that future research could investigate reducing the need for hypernetwork pretraining and further investigate the behaviour of hypernetwork-augmented language models.

\section*{Acknowledgements}

Research supported with Cloud TPUs from Google's TPU Research Cloud (TRC). We thank AllenNLP members and Jonas Pfeiffer for encouraging and useful coversations in earlier stages of this project.

\bibliography{anthology,custom}
\bibliographystyle{acl_natbib}

\appendix

\section{Dataset Details}

\subsection{Input Lengths}
\label{sec:median_calcs}
When calculating FLOPs estimates, we use the median sequence length of the inputs and outputs to calculate inference costs. We compute the median over the train split of Super-Natural Instructions and over 10,000 random samples from the T0 train split of P3. We calculate the medians for each format separately, rather than adding the instance and instruction-only values together (hence the mismatch in values). We provide the calculated values in Table~\ref{tab:medians}. We find that P3 inputs mostly consist of the instance, with prompt templates consisting of relatively few tokens, while SNI inputs consist mostly of instructions. This explains why HINT models are much cheaper than Tk-Instruct models, but not that much cheaper than T0 models, as HINT models reduce FLOPs by avoiding reprocessing the instruction with every input.

\begin{table}[]
    \centering
    \adjustbox{max width=.5\textwidth}{\begin{tabular}{@{}lcc@{}}
    \toprule
        & \multicolumn{2}{c}{\textbf{Median \# Tokens}} \\
        \cmidrule(r){2-3} \textbf{Text Sequence} & \textbf{SNI} & \textbf{P3} \\ \midrule
        Instance only & 44 & 81 \\
        Instruction only & 69 & 24 \\
        Instruction + Instance & 133 & 103 \\
        Instruction + 2 positives & 197 & - \\
        Instruction + 2 pos. + instance & 199 & - \\
        Output & 1 & 6 \\ \bottomrule
    \end{tabular}}
    \caption{Median sequence length, given in number of T5 tokens, for Super-Natural Instructions and P3.}
    \label{tab:medians}
\end{table}

\subsection{Split Sizes}

We report the sizes of splits here. For Super-Natural Instructions, we use the default setting from \citet{supernaturalinstructions} where 100 examples are provided for each task in train and test splits. We also note that we follow the sampling procedure used by \citet{sanh2022multitask}, where we ``treat
any dataset with over 500,000 examples as having 500,000 / \texttt{num templates examples}'' during training. Taking this sampling into account results in the much smaller dataset size seen in Table~\ref{tab:dataset_sizes}. We refer readers to \citet{sanh2022multitask} for more details on P3.

\begin{table}[]
    \centering
    \adjustbox{max width=.5\textwidth}{\begin{tabular}{@{}lcc@{}}
    \toprule
        \textbf{Dataset} & \textbf{Train} & \textbf{Test} \\ \midrule
        Super-Natural Instructions & 75,417 & 11,810 \\
        P3 & 90,897,454 & 2,940,068 \\
        P3 (adjusted for sampling) & 17,277,532 & 2,940,068 \\ \bottomrule
    \end{tabular}}
    \caption{Number of samples in given splits for each dataset.}
    \label{tab:dataset_sizes}
\end{table}

\section{Full P3 Results}
\label{app:p3_full}
We report the full results of models on the P3 dataset from Table~\ref{tab:p3_res} in Table~\ref{tab:p3_res_full}.

\begin{table*}
\centering
\adjustbox{max width=\textwidth}{
\begin{tabular}{lccccccccccc}
\toprule
 \textbf{Model} & \textbf{Avg} & \textbf{Rel. FLOPs} & \textbf{ANLI} & \textbf{HellaSwag} & \textbf{StoryCloze} & \textbf{CB} & \textbf{COPA} & \textbf{RTE} & \textbf{WiC} & \textbf{WSC} & \textbf{WinoGrande} \\\midrule
T0-3B & 54.9 & $\times$1.0 & 33.4 & 27.3 & 84.0 & 45.4 & 72.8 & 64.6 & 50.6 & \textbf{64.9} & 50.9 \\
T0-3B (our replication) & 64.4 & $\times$1.0 & \textbf{41.7} & 30.1 & \textbf{96.9} & 72.7 & 89.1 & 81.2 & 51.7 & 57.2 & 59.2 \\ 
T0-3B + PEFT & \textbf{65.5} & $\times$1.0 & 41.5 & 30.1 & 96.6 & \textbf{76.9} & \textbf{92.2} & 82.1 & \textbf{54.2} & 56.6 &  59.2 \\ 
No Prompt & 57.5 & \textbf{$\times$0.8} & 34.4 & 27.7 & 88.8 & 69.4 & 66.3 & 56.5 & 52.5 & 61.3 & 60.6 \\ \hdashline
Hypter (Joint) & 64.6 & $\times$1.0 & 41.1 & 29.4 & 96.7 & 76.3 & 87.4 & 79.6 & 52.1 & 58.3 & \textbf{60.9} \\
\textbf{HINT (Joint)} & 65.4 & $\times$1.1 & 41.6 & \textbf{30.3} & 96.6 & 76.0 & 88.8 & \textbf{84.2} & 51.4 & 59.5 & 60.1 \\
Hypter (Split) & 56.2 & \textbf{$\times$0.8} & 34.2 & 28.1 & 86.8 & 58.0 & 67.3 & 65.0 & 50.5 & 60.0 & 55.7  \\
\textbf{HINT (Split)} & 60.3 & \textbf{$\times$0.8} & 37.0 & 29.1 & 85.6 & 67.6 & 71.0 & 77.2 & 51.0 & 64.2 & 60.0 \\
\bottomrule
\end{tabular}}
\caption{Results on T0 evaluation tasks after training on the T0 P3 train set. We report averaged accuracy across prompts for each task, and the overall average performance (`avg'). FLOPs are calculated assuming we are processing 100 examples of a single task. The `Joint' and `Split' HINT variants refer to the two input formats for P3 described in Section~\ref{sec:exp_details}.}
\label{tab:p3_res_full}
\end{table*}

\section{Model Compute Calculations}

We provide a more thorough description of the compute and memory costs associated with various models we discuss here.

\subsection{Compute Costs}
\label{sec:compute_costs}
We will let $i$ be the sample length, $t$ be the task instruction length, $o$ be the output sequence length, $n$ be the number of same-task samples we wish to process, and $N$ be the number of parameters in the model. We will assume we are only processing examples from the same task.

\paragraph{Tk-Instruct} As Tk-Instruct concatenates instruction and sample together as input, it uses roughly $Nn(i + t + o)$ FLOPs.

\paragraph{HINT} The cost of the HINT model is more complicated. Let $N'$ be the cost of the hypernetwork generator, and $A$ be the cost of the parameter-efficient modules inserted into the underlying model. The cost of running the hypernetwork is $t(N + N')$ (since the hypernetwork encoder is the same size as the underlying model). The cost of then running the underlying model with parameter-efficient modules is $n(N+A)(i+o)$. We sum these two terms to get the total cost of HINT: $t(N + N') + n(N+A)(i+o)$. We do not consider the additional cost of inserting the instruction in the decoder as this only affects the few (usually 1-2) output tokens and the decoder cross-attention only, and so is negligible. We can simplify the HINT compute cost down by observing that in most cases $N >> A$ and $N >> N'$, resulting in the cost of HINT being roughly $tN + nN(i+o)$. This simpler formulation highlights the main benefit of HINT: the instruction no longer is processed with every sample, and so compute cost is $\propto t + n$ as opposed to $\propto tn$.

\subsection{Memory Costs}
\label{sec:memory_costs}
Here, we will let $l$ be the number of layers, $d$ the model hidden dimension, $h$ the number of heads, $k$ the size of the keys/values, and $s$ be the length of the sequence we want to save. We ignore bias terms for simplicity.

\paragraph{Decoder-only Models} If we want to cache the key/value pairs for a given sequence, we will store $2lhks$ values - a key and value for every head in every layer, for each item in the sequence. We note that typically $kh = d$ in models, and so in the main text we simplify this to $2lds \propto lds$.

\paragraph{HINT} In the default HINT setup, we save three elements: the processed instruction sequence, which contains $ds$ values (one vector per token); the adapter weights, $2*512ld$ values (one adapter comprising of two weight matrices per layer, where each weight matrix has size $512\times d$); the prefix values, $2*30lhk$ values (a 30-length prefix and key per layer per head) ). This sums to give a total memory cost of $ds + 1024ld + 60lhk$. Note that in the default HINT settings, we use prefixes of length 30 and adapters with bottleneck size 512, but these settings could be adjusted to reduce memory costs. Applying the simplification $kh = d$, we get that the HINT memory cost is $\propto ds + ld$.

\section{GPT-2 Instruction Finetuning}

When finetuning GPT-2 for Table~\ref{tab:sni_res}, we trained for [3, 5, 10] epochs with a batch size of 32. We use AdamW \citep{loshchilov2018decoupled} and swept learning rates of [$1\times10^{-5}$, $2\times10^{-5}$, $5\times10^{-5}$], using a linear warmup and decay schedule with 1,000 steps of warmup. We report the highest overall results in Table~\ref{tab:sni_res}. Following \citet{opt_iml}, we minimise the loss only over the target tokens (with EOS token added after the target answer), not the inputs, since these are always provided during test time.  Note that we calculate the FLOPs used by GPT-2 during inference based on the estimates provided by \citet{scaling_laws} that GPT-series models use $2N$ FLOPs per token, where $N$ is the number of parameters in the model.

\section{T0 Replication}
\label{sec:t0_rep}

\begin{figure*}
    \centering
    \adjustbox{max width=\textwidth}{
        \includegraphics{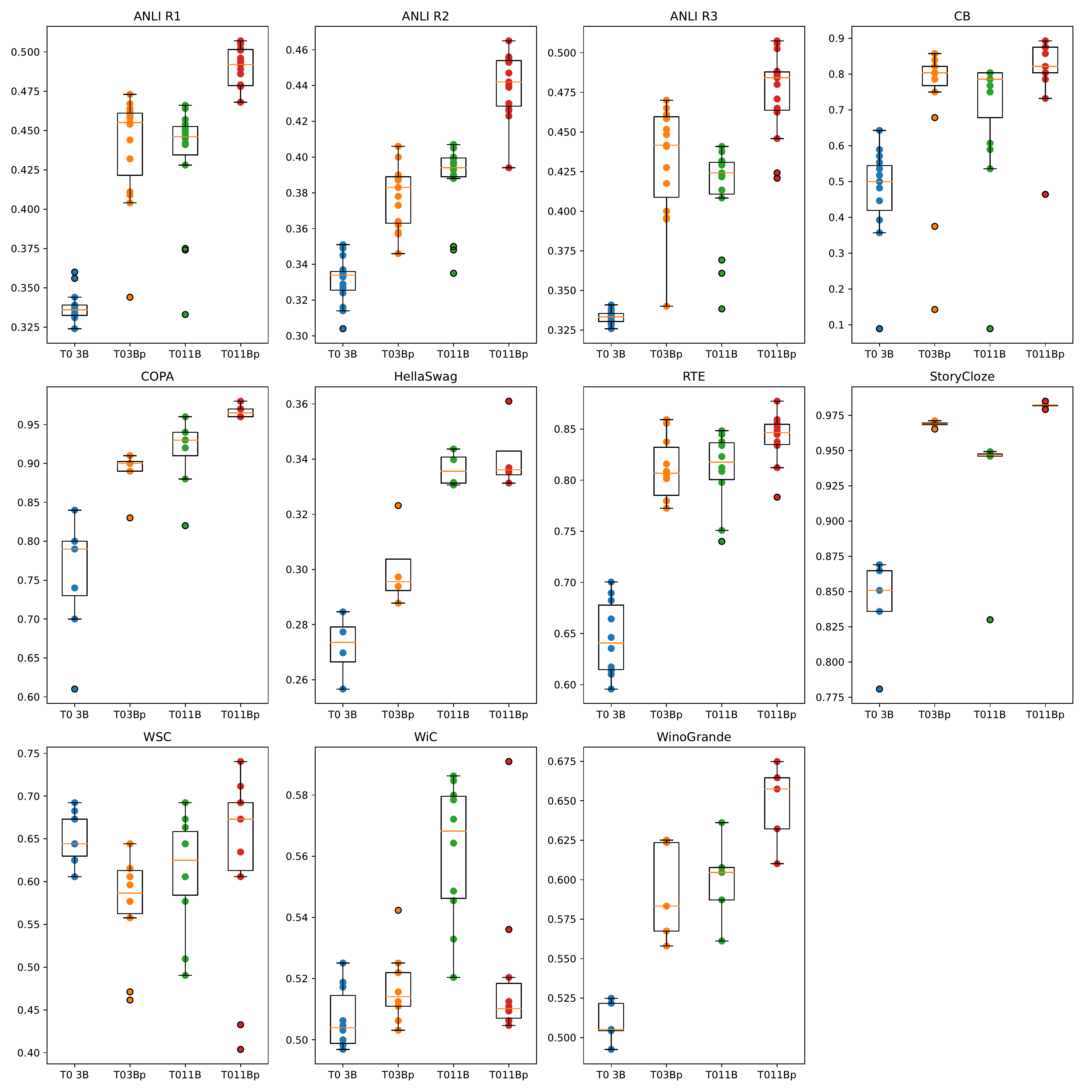}
    }
    \caption{T0 evaluation set accuracy across tasks. Each dot represents the performance from a different prompt. `T03Bp' and `T011Bp' refer to our T0 replications, while `T03B' and `T011B' are the models released by \citet{sanh2022multitask}.}
    \label{fig:p3_rep}
\end{figure*}

During initial experiments, we replicated the T0 training in the \texttt{t5x} framework, using the same training set and mixing proportions as \citet{sanh2022multitask}. We found that our replications performed significantly better than the reported T0 performance when trained for longer. We train 3B and 11B size models on the T0 training mixture for 20,000 steps using a batch size of 2048, a maximum input sequence length of 1024, a maximum output sequence length of 256, and the Adafactor optimizer with a constant learning rate of 0.001. We start from the T5 v1.1 + LM adaptation checkpoints and fully finetune the model. As seen in Table~\ref{tab:p3_rep}, our replications significantly outperform both T0 models, suggesting that T0 was undertrained. We also compare the variances in prompt performance in Figure~\ref{fig:p3_rep}.

\begin{table*}
\centering
\adjustbox{max width=\textwidth}{
\begin{tabular}{lcccccccccc}
\toprule
 \textbf{Model} & \textbf{ANLI} & \textbf{HellaSwag} & \textbf{StoryCloze} & \textbf{CB} & \textbf{COPA} & \textbf{RTE} & \textbf{WiC} & \textbf{WSC} & \textbf{WinoGrande} & \textbf{AVG} \\\midrule
T0-3B & 33.4 & 27.2 & 84.0 & 45.4 & 75.9 & 64.6 & 50.7 & \textbf{65.1} & 51.0 & 55.2 \\
T0-3B (ours) & \textbf{41.7} & \textbf{30.1} & \textbf{96.9} & \textbf{72.7} & \textbf{89.1} & \textbf{81.2} & \textbf{51.7} & 57.2 & \textbf{59.2} & \textbf{64.4} \\ \hdashline
T0-11B & 41.0 & 33.6 &  92.4 & 70.1 & 91.5 & 81.0 & \textbf{56.1} & 61.1 & 59.9 & 65.2 \\
T0-11B (ours) & \textbf{46.8} & \textbf{34.1} & \textbf{98.2} & \textbf{81.2} & \textbf{96.6} & \textbf{84.0} & 52.1 & \textbf{62.6} & \textbf{64.8} & \textbf{68.9} \\
\bottomrule
\end{tabular}}
\caption{T0 evaluation task accuracy, comparing results using models from \citet{sanh2022multitask} and our own replications. We report accuracy averaged across prompts for the same dataset. For ANLI, we first average performance across prompts for ANLI R1/2/3 separately, and then average the three results.}
\label{tab:p3_rep}
\end{table*}

\end{document}